# PLOT 94

## in ambiance X-WINDOW[*]

I. Vega-Páez and C. Hernández-Hernández [1]

e-mail:
vegapaez@mvax1.red.cinvestav.mx
chernand@mvax1.red.cinvestav.mx

**Department of Electric Engineering
Section Compute
Center of Investigation and Advanced Studies of the
Polytechnic National Institute,
México, D.F.**

## Summary

<PLOT > is a collection of routines to draw surfaces, contours and so on. In this work we are presenting a version, that functions over work stations with the operative system UNIX, that count with the graphic ambiance X-WINDOW with the tools XLIB and OSF/MOTIF. This implant was realized for the work stations DEC 5000-200, DEC IPX, and DEC ALFA of the CINVESTAV (Center of Investigation and Advanced Studies). Also implanted in SILICON GRAPHICS of the CENAC (National Center of Calculation of the Polytechnic National Institute ).

## Keywords

Plot; draw Surfaces; contours; differential equations; visualizing, X-WINDOW, OSF/MOTIF.

## Record

The group of routines <PLOT> was originally developed to draw over the plotter Calcomp in the decade of the 70s by teacher Harold V. McIntosh in Institute National of Nuclear Engineering (McIntosh, 1975), this was result of the interest to visualize functions that were solutions of problems so as practical and theorists, such as the differential equations. The enthusiasm to see the results drawn carry to <PLOT> to be implanted in distinct centers of investigation as the university of Utah by Teacher Nelson H. Beebe <PLOT 79> (Beebe, 1979). All the implantations were realized in language FORTRAN until the middle of the 80's, between the implantations in FORTRAN that followed to <PLOT 79> are: <PLOT 84> that functions over the operative system CP/M-8080 realized by Francisco Gil (Gil and McIntosh, 1984), also functions over the computer IBM 1130 of the University of Puebla. With the apparition of the language "C" a version in this language was initiated in 1990 by the Teacher Gerardo Cisneros S. and Angel Bernardo Canto G. That finally remained finished in 1993 <PLOT 90> (Canto, 1993), this version remained as Library alternates to written in Fortran for the PC-Compatible reported by José P. Hernández E. (Hernández, 1993).

[1] Members of the Section of Computing of the CINVESTAV (Center of Investigation and Advanced Studies of the Polytechnic National Institute, México)
* This work was supported in part by Council National of Science and Technology (Consejo Nacional de Ciencia y Tecnología CONACYT).

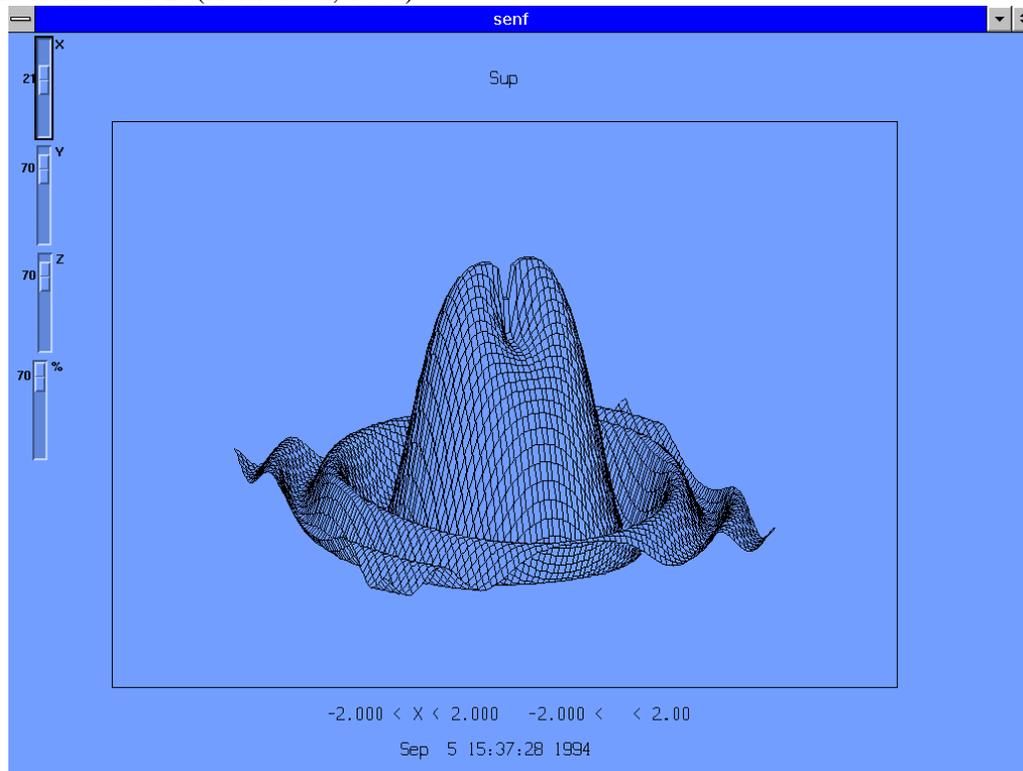

**Figure 1.** Draw surface of rectangular rank

With the purchase of work stations by the section of Computing of the CINVESTAV the version <PLOT 94> was development based on the version <PLOT 90c> (Canto, 1994). For our version we are using the interface graphics X-WINDOW with the tools XLIB and OSF/MOTIF, tells interface was elected because it has been converted practically in the standard Graphic of the computers that have the operative system UNIX (Reiss and Radin, 1992). On the other hand, the program <GEOM> developed originally by teacher Harold V. McIntosh in Superior School of Physics and Mathematics of the I.P.N. (Polytechnic National Institute) to draw based molecules in primitive <PLOT>, continued the development in Institute of Sciences of University of Puebla where developed <GEOM> in language "C" in two versions: A for the PC-compatibles that was used to study the lunar shade projected over the earth in the memorable total eclipse of sun in June, 1991 (McIntosh, 1993a), the other version was developed for C-objective of the operative system of NextStep (McIntosh, 1993b), in this last version of <GEOM> was where our work was inspired. The figures [1], [2] y [3] exemplified the work realized in the control of the angles of Euler and the parameter of resolution thus as the simultaneous display of varied demonstrations.

**Methodology**

The implantations of <PLOT> mentioned with priority, generally were designed to be used no friendly way, this due to the limitations of hardware existing at the moment. Currently exist sufficiently rapid computers to which been graphic user interfaces (GUI) incorporated which permit the interaction with the computer of friendly way, in this scope it is important to integrate <PLOT> to the ambiance of a graphic interface.

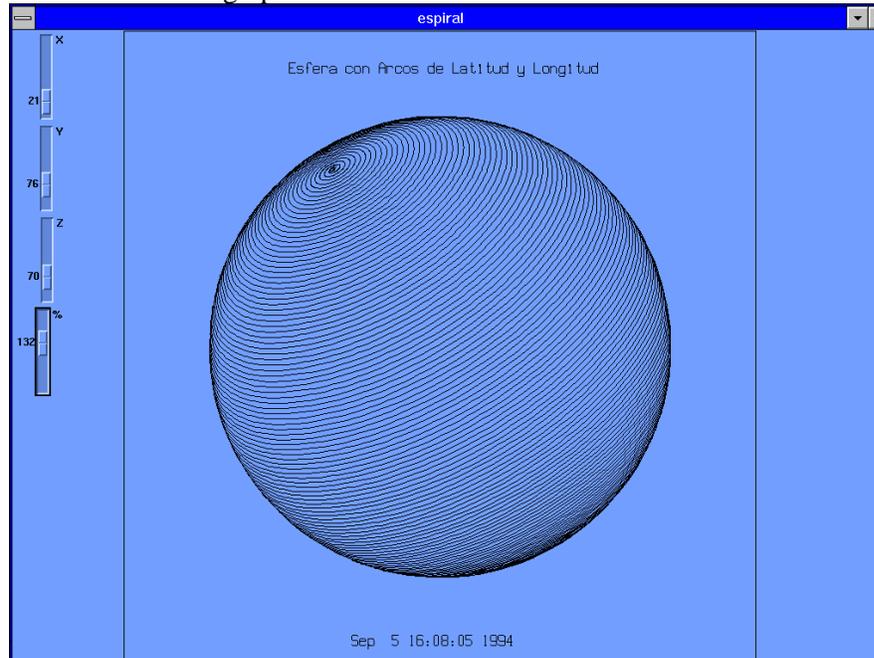

**Figure 2.** Draw closed surface

One of the important aspects to be not to alter substantially the specifications of use standard of <PLOT> inside of the graphic interfaces, in other words a habitual user of <PLOT> will not have many difficulties to use <PLOT94>, of equal manner to use this version follow effective the specifications of use most some additions. To realize the work plant two phase of work.

## FIRST PHASE

In first instance modified the routines that involve to the device of as graphic such display:

*plot.c, ploti.c, plotsal.c plt00.c point.c and so on.*

Also added the routine *plomth.c* that gives the possibility of having the exit to the plotter HP-DraftPro Plus with the graphic language HP-GL.

Before having the routine *plomth.c* realized the program *hitohp.cnv* written in language CONVERT (Guzman, 1966) and (McIntosh and Cisneros, 1990) that realized the conversion of graphic language of the plotter Hiplot DMP-29 of Houston Instrument to the graphic language HP-GL.

## SECOND PHASE

As second part of the work a group of routines in charge of managing the graphic interfaces were developed thus as the execution of the routines of <PLOT>. This scheme was realized by means of The calls to XLIB and OSF/MOTIF, and the routine *plot94.c* respectively.

For the use of <PLOT94>, in every case, to be believed the routine *plot94.c* that consists of the code as in <PLOT90c>, removing the routines *pltdk.c* and *closeplt.c* that are included inside of the routines of control, furthermore of the adequate additions for the functioning of X-WINDOW.

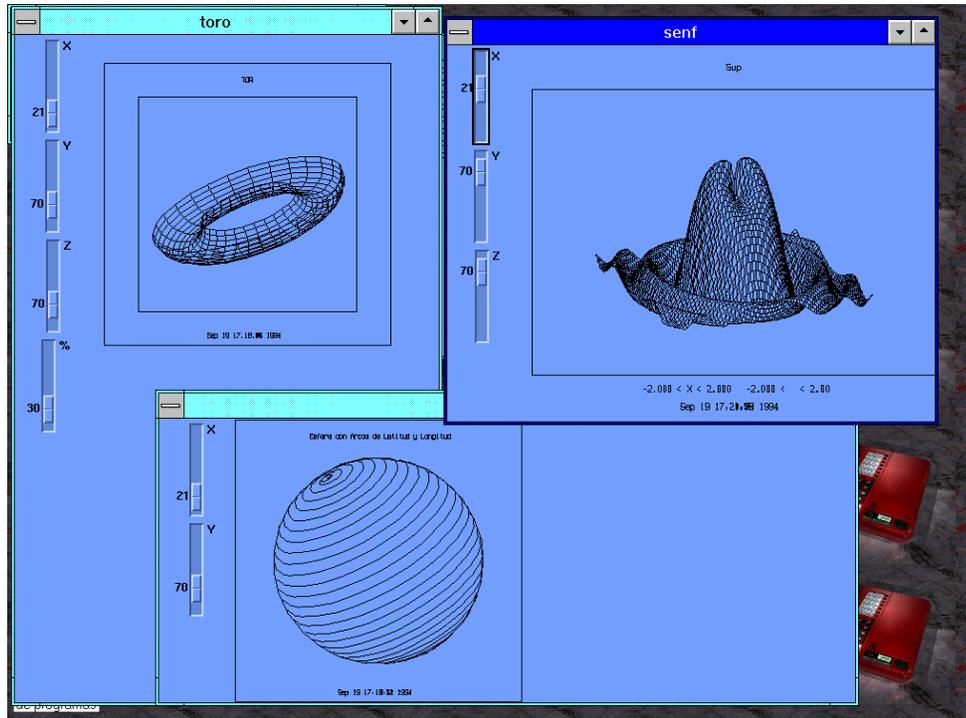

**Figure 3.** Display simultaneous of draw surfaces

## Conclusion

This to implant by the rapidity of calculation in the current computers, offers a visualization of functions that were solutions of problems so as practical and theorists, such as the differential equations, with that can achieve a three-dimensional sensation, with a specification of use easy.